\documentclass{article}

\usepackage[nonatbib,final]{neurips_2018}

\usepackage[utf8]{inputenc} 
\usepackage[T1]{fontenc}    
\usepackage{hyperref}       
\usepackage{url}            
\usepackage{booktabs}       
\usepackage{amsfonts}       
\usepackage{nicefrac}       
\usepackage{microtype}      
\usepackage{graphicx}
\usepackage{subfigure}
\usepackage{helvet}
\usepackage{courier}
\usepackage{amsmath}
\usepackage{amssymb}
\usepackage{tabularx}
\usepackage{xcolor}
\usepackage{graphicx}
\usepackage{bbm}
\usepackage{bm}
\usepackage{colortbl}
\usepackage[clock]{ifsym}
\usepackage{enumitem}
\usepackage{multicol} 
\usepackage[utf8]{inputenc}
\usepackage{booktabs}  
\usepackage{algorithm}
\usepackage{algorithmic}
\usepackage{wrapfig}
\usepackage{amsthm}
\usepackage{multirow}
\usepackage{setspace}

\newcolumntype{C}[1]{>{\centering\arraybackslash}p{#1}}
\newcolumntype{R}[1]{>{\raggedright\arraybackslash}p{#1}}

\title{A Bayesian Approach to Modelling Longitudinal Data in Electronic Health Records}

\author{
  Alexis Bellot$^{1,2}$\hspace{0.5cm} Mihaela van der Schaar$^{1,2,3}$\\
  $^{1}$University of Cambridge, $^{2}$The Alan Turing Institute, $^{3}$University of California Los Angeles\\
  \texttt{[abellot,mschaar]@turing.ac.uk} \\
}

\begin{document}

\maketitle

\begin{abstract}
Analyzing electronic health records (EHR) poses significant challenges because often few samples are available describing a patient's health and, when available, their information content is highly diverse. The problem we consider is how to integrate sparsely sampled longitudinal data, missing measurements informative of the underlying health status and fixed demographic information to produce estimated survival distributions updated through a patient's follow up. We propose a nonparametric probabilistic model that generates survival trajectories from an ensemble of Bayesian trees that learns variable interactions over time without specifying beforehand the longitudinal process. We show performance improvements on Primary Biliary Cirrhosis patient data.
\end{abstract}

\section{Introduction}
\begin{figure*}[t]
\centering
\includegraphics[width=1\textwidth]{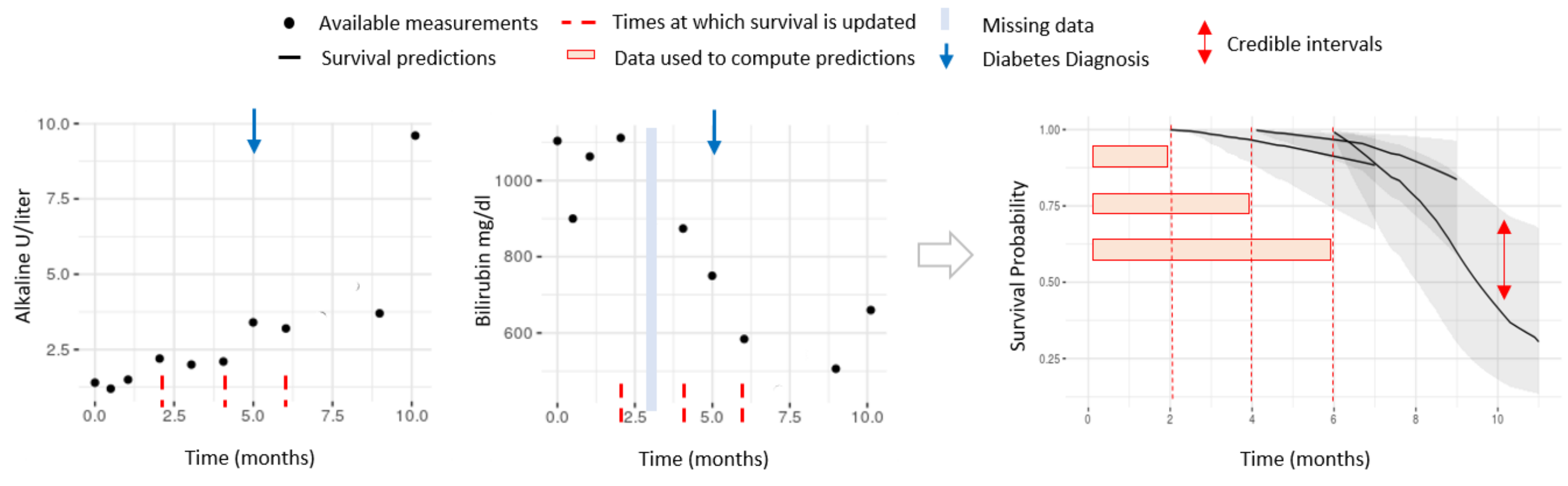}
\caption{Illustration of the available patient's record and subsequent model survival predictions for an exemplary patient with primary biliary cirrhosis. The left and middle panel suggest that the patient gradually transitions from being healthy to becoming sick as elevations in bilirubin measurements over time signal an advanced disease stage; a fact reflected in our model's survival predictions (right panel). The times at which survival predictions are updated are given by the red dashed lines.}
\label{surv_curves2}
\end{figure*}

Clinical prognostic models derived from electronic medical records are an important support for many critical diagnostic and therapeutic decisions. The majority of these models, however, do not leverage the information contained in a patient's history, such as tests, vital signs, and biomarkers. Longitudinal EHR data is increasingly relevant to better understand and predict the health state of patients \textcolor{blue}{\cite{stanziano2010review}} but remains a challenge for analysis because of irregular, sporadic sampling and high missingness (often informatively) \textcolor{blue}{\cite{ferrer2017individual}}. As an \textbf{example} of a typical trajectory found in EHRs we depict in Figure \textcolor{blue}{\ref{surv_curves2}} two longitudinal variables describing a patient suffering from liver disease.

In this paper we consider the problem of \textit{analyzing survival of patients on the basis of scarce longitudinal data sampled irregularly over time}. Modelling sparsely sampled data - arising when intervals between observations are often large - is becoming increasingly relevant for managing elderly patient populations due to the greater prevalence of chronic diseases that develop over a period of years or decades \textcolor{blue}{\cite{licher2019lifetime,wills2011life}}. A substantial portion of machine learning research now investigates prognosis with time series data, typically focusing on patients in the hospital where information is densely collected. Predictions in this setting tend to target a binary event and require structured and regularly sampled data \textcolor{blue}{\cite{choi2016doctor,lipton2017doctor}}. Modelling survival data differs from the above as data is often recorded with censoring - a patient may drop out of a study but information up to that time is known. Moreover, the sparsity of recorded information will tend to lead to substantial uncertainty about model predictions which needs to be captured for reliable inference in practice \textcolor{blue}{\cite{begoli2019need}}. We discuss these issues and develop a Bayesian nonparametric model that aims to overcome these difficulties by \textbf{(1)} quantifying the uncertainty around model predictions in a principled manner at the individual level, while \textbf{(2)} avoiding to make assumptions on the data generating process and adapting model complexity to the structure of the data. Both contributions are particularly important for personalizing health care decisions as predictions may be uncertain due to lack of data, while we can expect the underlying heterogeneous physiological time series to vary wildly across patients.


\section{Problem Formulation}
\label{problem_formulation}
Our goal is to analyze the occurrence of the event of interest in a medically-relevant time frame $(t, t + \Delta t]$. The survival probability given survival up to time $t$ is given by,
\begin{align*}
    S(t+\Delta t; t) := \mathbb P(T>t + \Delta t |T>t)
\end{align*}
Each individual $i$ in our population is described by an irregularly sampled time series, defined as a sequence $\mathbf X_i := (X_{i1}, ... , X_{i|\mathbf t_i|})$ of multidimensional observations (in $\mathbb R^d$) with irregular intervals between their observation times $\mathbf t_i = (t_{1}, ... , t_{n_i})$. 
We denote by $\mathbf X_i(s)$ all information available for patient $i$ up to time $s$. Our target for prediction is the time until the occurrence of the event of interest $T_i\in\mathbb R^+$. For some patients this outcome may be censored (with censoring time denoted $C_i$), for example if they drop out of the study. We write $\delta_i = 1$ for patients experiencing the event of interest and $\delta_i=0$ for censored patients. Let $\mathcal{D}_t$ denote $n$ independent samples of the random tuple $\{\mathbf X_i(t),\delta_i,\delta_i T_i + (1-\delta_i)C_i\}$ on which a data-driven model is to be fitted. 

Note that it will also be useful to define missing data as those missing entries in the multidimensional feature vector $X_{it}$, $t \in \mathbf t_i$. This is because when other information is observed at a time $t$, missing risk factors may indicate a perception of reduced relevance in certain patients, as is the case with routine measurements such as the Body Mass Index \textcolor{blue}{\cite{bhaskaran2013representativeness}}. 

\section{Model description}
In this section we introduce our proposed modelling approach, termed Bayesian Nonparametric Dynamic Survival (\texttt{BNDS}) model.

We represent a patient trajectory explicitly as a discrete sequence of tuples consisting of the observation times and corresponding measurements $(\mathbf t_i, \mathbf x_{i})$. For each one of these sequences, we define event indicators $\mathbf w_i \in \{0,1\}^{|\mathbf t_i|}$, which represent the survival status of a patient $i$ (i.e. $w_{it_{ij}}=0$ if the patient was alive at time $t_{ij}$). The probability of an event is described by $p(w_{it_{ij}}=1) = p(T_i \leq t_{ij}|T_i>t_{ij-1})$.

This interpretation as conditional probabilities is especially useful to define full survival distributions. Define a fine sequence of time windows $(t_k,t_{k+1})_{r=k}^{r=s}$ in the future. The survival probability at a time $t_s$ in the future given survival up to time $t_k$ then is given by \textcolor{blue}{\cite{singer1993s}},
\begin{align}
   p(T_i>t_{s} |T_i>t_{k})  = \prod_{r=k+1}^s p(T_i>t_r|T_i>t_{r-1})  = \prod_{r=k+1}^s (1-p(w_{it_r} = 1))
\end{align}

\textbf{Prior model.} The nonparametric nature of inference follows from first assigning a prior $\Pi$ over a random basis of regression trees $h_k$ that captures the interaction of time $t_r$ and the values of the $i^{th}$ patient variables at that time, denoted by $\mathbf{x}_{it_r}$\footnote{The prior on the trees itself, $\Pi:= (\mathbf Z,\mathbf M) = (Z_k, M_k)_k$ is composed of priors on the tree structure $\mathbf Z$ (depth of the tree and, splitting variables and values in each node) and leaf node outputs $\mathbf M$. We refer the reader to \textcolor{blue}{\cite{chipman2010bart}} and the Supplement for more details.}. For a given prior $\Pi$, the generative model proceeds as follows: we sample $k$ regression trees, combine them to form an ensemble and transform their output to issue probabilities $p(w_{it_r}=1)$ for each tuple $(t,x_{it_{r-1}})$, $t_r \in \mathbf t_i$. In a nutshell,
\begin{align}
    \mathbf h \sim \Pi, \quad    p_{it_r}|\mathbf h, t,\mathbf x_{it_{r-1})} \sim \Phi \left(\sum_k h_k(\mathbf x_{i_{r-1}},t) \right), \quad
    W_{it_r}|p_{it_r} \sim \text{Bern}(p_{it_r})
\end{align}
$\Phi$ denotes the cumulative distribution function of a standard Gaussian random variable. In the medical setting this set up is useful as prior beliefs on the importance of risk factors for a disease, such as diabetes, can be included in $\Pi$ to encourage trees to split on that variable or to regulate the depth of the trees based on the believed complexity of associations. Refer to the Supplement for more details.

\textbf{Posterior distribution.}
Our observation model guides our probability model to a posterior distribution of parameters that agrees with the observed data. It is defined as follows,
\begin{align}
\label{likelihood}
p(\mathbf w) = \prod_{i}p(\mathbf w_i) = \prod_{i}\prod_{j=1}^{|\mathbf t_i|} p(w_{it_j}=1)^{w_{it_j}}\left( 1 - p(w_{it_j}=1) \right) ^{1-w_{it_j}}
\end{align}
We compute the posterior $p(\mathbf h | (\mathbf w_i)_i)$ (intractable because of the large parameter space) of the set of tree structures and leaf node parameters $(\mathbf M,\mathbf Z)$ given $(\mathbf w_i)_{i=1}^N$ via repeatedly sampling from a tailored version of the Backfitting MCMC algorithm introduced in \textcolor{blue}{\cite{chipman2010bart}}. In a given iteration, we proceed by subtracting from the observed $\mathbf w_i$'s the fit of the current ensemble and draw the subsequent tree conditional on the resulting residual with a Gibbs sampler and a Metropolis step. Because we target binary outcomes, we use a data augmentation approach \textcolor{blue}{\cite{albert1993bayesian}} which introduces $i.i.d.$ Gaussian latent variables $V_{it_{ij}}$, $i=1, ..., N$, (one for each observed patient and observation time) simulated in an additional step in the MCMC algorithm. Let $f_{i,t_k}:=h_k(\mathbf x_{it_{k-1}},t_k)$. $V_{it_k}$ is generated conditional on the observed data $w_{it_k}$ as follows,
\begin{align*}
    V_{it_k} |  w_{it_k}=1 &\sim \text{max}\left(\mathcal{N} \left(f_{i,t_k}, 1 \right),0 \right), \quad
    V_{it_k} |  w_{it_{k}}=0 \sim \text{min}\left(\mathcal{N} \left(f_{i,t_k}, 1 \right),0 \right)
\end{align*}
Draws from this posterior are used to sample real-valued latent variables that link the probit model for binary outcomes with the tree regression model. This step can be inferred exactly with a Gibbs sampler as the posterior has a well-defined closed form \textcolor{blue}{\cite{kapelner2013bartmachine}}. For an individual $i$ having survived up to $t_k$, the survival probability beyond time $t_{k+s}$ is estimated by extrapolating along the time dimension on this surface,
\begin{align}
    p(T_i>t_{k+s}|T_i>t_k, \mathcal D_{t_k}):=\prod_{r=k+1}^s (1-p(w_{it_r}=1|\mathcal D_{t_k})),
    \label{posterior_survival}
\end{align}
$\mathcal D_{t_k}:=\{\mathbf X_i(t_k)\}_{i=1}^N$ is all patient information up to time $t_k$. We quantify uncertainty around model predictions by considering the quantiles of the posterior distribution, providing credible intervals, which are especially important for heterogeneous data samples as flexible algorithms might overfit the training data and give unreliable predictions at test time.

\textbf{Informative missing values.}
As we have noted, missing measurements may often reflect decisions by clinicians which implicitly provide information about the health status of the patient \footnote{This setting corresponds to data missing not at random (MNAR), see \textcolor{blue}{\cite{little2014statistical}} for more details.}. To include this information, we augment the space of splitting rules in the stochastic tree construction to include missingness of a variable as a splitting criterion similarly to the approach of \textcolor{blue}{\cite{twala2008good,kapelner2013bartmachine}}. Such a split, if informative for survival (in the sense of increasing the likelihood of observed event times under the current model) will be encouraged by the MCMC sampling algorithm and result in meaningful partitions of the data based on unobserved measurements. For a possible splitting threshold $\delta$ in the range of a patient variable $z$ with missing values, three splitting rules are considered:
\textbf{(1)} - $\{i: z_i \leq \delta \text{ or } z_i \text{ missing }\}$ versus $\{i: z_i > \delta \}$, \textbf{(2)}
- $\{i: z_i \leq \delta \}$ versus $\{i: z_i > \delta \text{ or } z_i \text{ missing }\}$, and \textbf{(3)}
- $\{i: z_i \text{ missing }\}$ versus $\{i: z_i \text{ not missing }\}$.

These are uniformly sampled as a potential splitting rule in the MCMC procedure. No imputation of training or testing data is required. 

\section{Experiments: Predictive performance comparisons}

\textbf{Data description.} Primary Biliary Cirrhosis (PBC) is a slowly progressing disease of the liver that affects approximately $0.025\%$ of the population, the majority of which are women. Early identification is difficult because many biomarkers are involved in its development with poorly understood interactions \textcolor{blue}{\cite{lindor2009primary}}. We considered data from the Mayo Clinic trial used previously in \textcolor{blue}{\cite{therneau2000extending}}. A total of $312$ patients were analyzed in this study of which $63\%$ experienced liver failure, the end point of interest. On average $6$ follow-up visits were recorded during an average time to event of $6.2$ years. Summary statistics are included in the Supplement.

\textbf{Baseline prediction algorithms.} Modelling longitudinal data to inform survival has been considered in two statistics approaches: Joint models \textcolor{blue}{\cite{andrinopoulou2012introduction,rizopoulos2011dynamic,hickey2016joint}} and Landmarking \textcolor{blue}{\cite{van2008dynamic,van2007dynamic,van2011dynamic}}. Joint models (JM) specify separate longitudinal and survival models related by shared random effects, inducing correlation between the two and propagating the uncertainty from future longitudinal estimates into survival predictions. Landmarking, in contrast, uses the last longitudinal measurement carried forward as an estimate for the current longitudinal value and builds a (static) prediction model based on these values. Its implementation consists of two steps: the construction of a cross-sectional data set, and the development of a Cox model based on that landmark data set. We evaluate joint models with the implementation by \textcolor{blue}{\cite{hickey2016joint}}. We used a multivariate normal linear mixed model for longitudinal variables and a Cox model to relate these and fixed variables to survival estimates. We implemented two variants of the landmarking approach: one using a traditional Cox survival model \textcolor{blue}{\cite{Cox:72}} and one using Random Survival Forests (RSF) \textcolor{blue}{\cite{ishwaran2008random}}. We discuss other related work in the Supplement.

\textbf{Performance evaluation.} We use the time-dependent concordance index ($C$-index) \textcolor{blue}{\cite{gerds2013estimating}} to assess the discriminative ability of all models, 
\begin{align}
    C(t, \Delta t) := \mathbb P \left(\hat{S}_i(t+\Delta t; t) > \hat{S}_j(t+\Delta t; t) | \delta_i=1, t\leq T_j, T_i>T_j \right)
    \label{cindex}
\end{align} 
The $C$-index corresponds to the probability that predicted time-to-event probabilities at a time of interest $t+\Delta t$ are ranked in \textit{accordance} to the actual observed survival times (at time $t+\Delta t$), but evaluated only for those patients still alive at the follow up time $t$ at which predictions are made. It thus serves as a measure of a model's discriminative power. The $C$-index is bounded between $0.5$ and $1$, $0.5$ indicating performance of random guesses and $1$ indicating perfect ordering of event times. In all experiments we implemented \texttt{BNDS} with $50$ trees and $1000$ samples after a burn-in of $250$ samples, our default specification.

\textbf{Results.} We observe in Table \textcolor{blue}{\ref{table:performance}} that \texttt{BNDS} significantly outperforms all other benchmarks. \texttt{BNDS} is the only method modelling longitudinal data agnostically and using the whole patient history. Both Cox-Landmark and JM require specification of variable interactions which strongly limits the structure they are able to recover from data. This limitation is further illustrated by the performance gain of RSF-Landmark over Cox-Landmark, arising because the former models variable interactions more flexibly. Notice also that the performance gap between JM and \texttt{BNDS} (both using the whole patient history) widens for predictions at later follow ups with respect to cross-sectional models, thereby highlighting the importance of considering a patient's history as well as current measurements.

\begin{table*}[t]
\fontsize{8}{8.5}\selectfont
\centering
\begin{tabular}{ p{2cm}p{1.2cm}C{1.7cm}p{1.2cm}C{1.7cm}p{1.2cm}C{1.7cm}  }
 \multicolumn {2}{ c }{\textbf{C-index on  PBC data}} & & & & & \\
  \arrayrulecolor{black}\midrule\\[-1ex]
Cox-Landmark  & $At$ $t=2y$ & $0.845\pm0.010$ &  $At$ $t=4y$ & $0.738\pm0.056$ &  $At$ $t=6y$ & $0.718\pm0.031$ \\
RSF-Landmark  &             & $0.873\pm0.008$ &              & $0.790\pm0.046$ &              & $0.772\pm0.032$ \\
Joint Model   &             & $0.877\pm0.009$ &              & $0.791\pm0.041$ &              & $0.787\pm0.031$ \\
BNDS (ours)   &             & $0.894\pm0.010$ &         & $0.809\pm0.043$  &        & $0.805\pm0.030$ \\[0.5ex]
\end{tabular}
\caption{Median $C$-index (higher better) and standard deviation using 5-fold cross-validation.}
\label{table:performance}
\end{table*}

\section{Conclusion}
We have introduced a Bayesian nonparametric method that is able to use sparse, longitudinal data to give personalized survival predictions that are updated as new information is recorded. Our modelling approach has the advantage of providing uncertainty estimates and has the flexibility to model highly heterogeneous data without a priori modelling choices. 

\bibliographystyle{plain}
\bibliography{bibliography}

\end{document}